\documentclass{article}

\PassOptionsToPackage{numbers, sort&compress}{natbib}

    \usepackage[preprint]{neurips_2021}

\usepackage[utf8]{inputenc} 
\usepackage[T1]{fontenc}    
\usepackage{hyperref}       
\usepackage{url}            
\usepackage{booktabs}       
\usepackage{amsfonts}       
\usepackage{nicefrac}       
\usepackage{microtype}      
\usepackage{xcolor}         

\usepackage{graphicx}
\usepackage{multirow}
\usepackage{multicol}
\usepackage{amsmath}
\usepackage{wrapfig,lipsum}
\usepackage{enumerate}
\usepackage{tikz}
\usepackage{pifont}
\newcommand{\etal}{\textit{et al.}}
\newsavebox{\tempbox}

\makeatletter
\newcommand{\printfnsymbol}[1]{%
  \textsuperscript{\@fnsymbol{#1}}%
}

\title{Towards Brain Inspired Design for Addressing the Shortcomings of ANNs}

\author{Fahad Sarfraz\printfnsymbol{1}, Elahe Arani\thanks{Contributed equally.} \& Bahram Zonooz \\
\textsuperscript{\rm 1}Advanced Research Lab, NavInfo Europe, Netherlands\\
\textsuperscript{\rm 2}Dep. of Mathematics and Computer Science, Eindhoven University of Technology, Netherlands\\
\texttt{\{f.sarfraz, e.arani, b.zonooz\}@tue.nl} \\
}

\begin{document}

\maketitle



\begin{abstract}
As our understanding of the mechanisms of brain function is enhanced, the value of insights gained from neuroscience to the development of AI algorithms deserves further consideration. Here, we draw parallels with an existing tree-based ANN architecture and a recent neuroscience study~\cite{shadmehr2020population} arguing that the error-based organization of neurons in the cerebellum that share a preference for a personalized view of the entire error space, may account for several desirable features of behavior and learning. We then analyze the learning behavior and characteristics of the model under varying scenarios to gauge the potential benefits of a similar mechanism in ANN. Our empirical results suggest that having separate populations of neurons with personalized error views can enable efficient learning under class imbalance and limited data, and reduce the susceptibility to unintended shortcut strategies, leading to improved generalization. This work highlights the potential of translating the learning machinery of the brain into the design of a new generation of ANNs and provides further credence to the argument that biologically inspired AI may hold the key to overcoming the shortcomings of ANNs.

\end{abstract}

\section{Introduction}
\label{sec:intro}


Artificial neural networks (ANNs) have achieved remarkable performance on many vision tasks which have been enabled by the considerable progress in developing deeper and more complex network architectures. However, despite the performance gains, the existing networks have been shown to be brittle and have several limitations and shortcomings. They require huge amounts of data to train, struggle with noisy and imbalanced datasets, do not generalize well to out-of-distribution data, and are vulnerable to shortcut learning and adversarial attacks~\cite{vandervert2016prominent}. While there have been studies on addressing these challenges individually, the majority of these specialized techniques and regularization approaches for overcoming a specific challenge lead to a trade-off in performance and do not provide a general solution~\cite{tsipras2019robustness}.

Humans, on the other hand, excel at learning efficiently even under challenging scenarios with limited data and can generalize well to novel scenarios.
Neuroscience has made substantial progress in understanding the mechanisms of brain functions and the design principles it employs to enable efficient learning~\cite{hassabis2017neuroscience,kudithipudi2022biological,hawkins2021thousand,macpherson2021natural}. It is, therefore, important to further exploit insights from our enhanced understanding of the learning machinery of the brain into the development of AI algorithms.



\begin{figure*} 
  \begin{center}
  \includegraphics[trim={0 0 1.5cm 0},clip, width=1\textwidth]{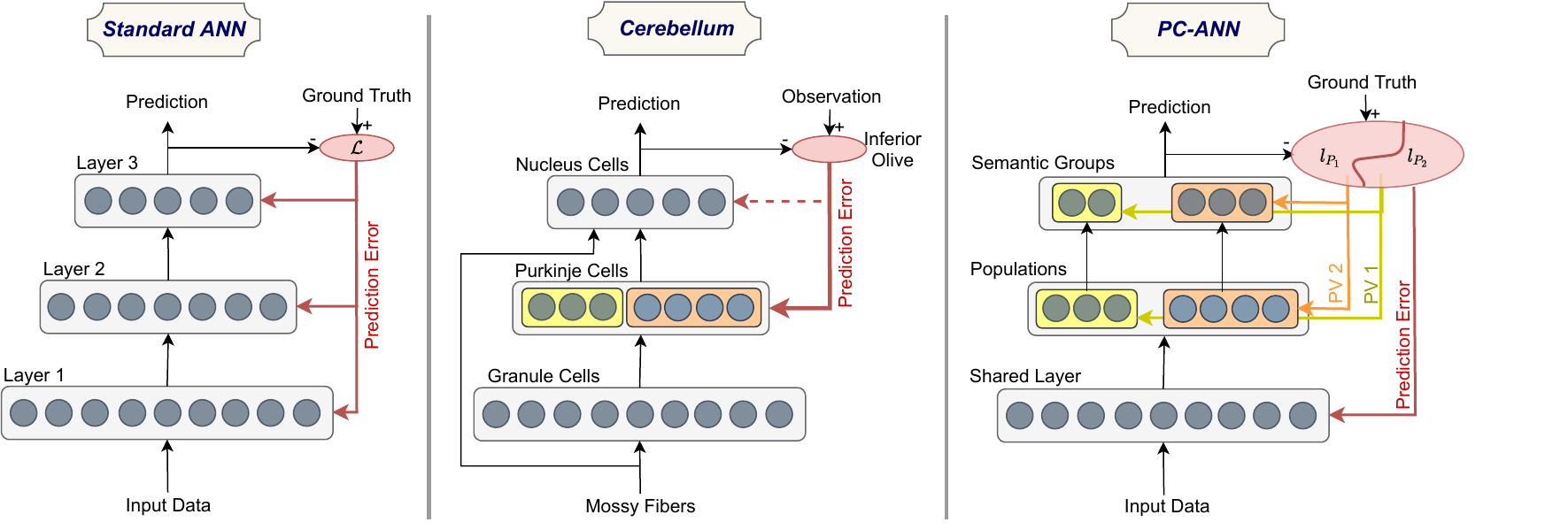}
  \caption{In \textit{standard ANN}, there is no grouping of neurons within a layer, and the predictions from Layer 3 are compared to the ground truths and then the overall prediction error signal is fed back to every neuron in the network which adjusts their weights to minimize the loss. 
  The \textit{Cerebellum} in the brain, on the other hand, has a vastly different design and learning mechanism. It resembles a three-layer feedforward network comprising granule cells as the input layer, Purkinje cells in the middle layer, and deep cerebellar nucleus (DCN) cells as the output layer. The predictions of each DCN neuron are compared to the actual observation, resulting in an error signal originating in the inferior olive. Unlike in standard ANN, the olive organizes Purkinje cells in the hidden layer into small populations which receive a limited personalized view of the error space. 
  Similar to the cerebellum, the desired population coding based ANN (\textit{PC-ANN}) would form an error-dependent grouping of neurons into populations that learn from partial error views from the classification error associated with learned semantic groupings of classes. We consider SplitNet as an instance of an architecture that bears some similarities to the desired population coding based architecture.
  }
   \label{fig:overview}
   \end{center}
   \vskip -0.2in
\end{figure*}

We consider a recent study~\cite{shadmehr2020population} which examines the organization of neurons in the cerebellum, an important learning site in the brain and resembles a three-layer feedforward network (see Figure \ref{fig:overview}). The neurons in the middle layer of the cerebellum are grouped into small populations which receive a personalized view of the entire error space. This is in stark contrast to standard ANNs which lack any such organization of neurons and each unit in the network receives the same error signal. We, thus, attempt to study the potential effect of a similar error-dependent organization of neurons in ANNs. To this end, in the object recognition task in ANNs, we consider the classification error associated with a learned semantic grouping of object classes as partial views of the error space and the corresponding set of disjoint subnetworks as populations that share a preference for a particular partial error view. From this perspective, our intended learning paradigm can be more aligned with tree-structured ANNs.

Thus, we consider SplitNet~\cite{kim2017splitnet}, originally proposed to improve inference speed and reduce the number of parameters, as a suitable tree-structured method to assess the potential of population coding in ANNs as we see many similarities between its network design and the error-dependent organization of neurons in the cerebellum (see Figure \ref{fig:overview}). Analogous to the grouping of neurons in the cerebellum, SplitNet learns to split the network weights into multiple groups that use disjoint sets of features.
In particular, since the logit values associated with the semantically disparate class groups only depend on the corresponding branched subtrees of the network and not the other subtrees, each group (subtrees) receives a gradient signal which is biased towards correcting the error associated with their corresponding semantic group (partial error view), similar to how populations in the cerebellum share preference for a biased error view.
Finally, similar to the cerebellum which receives a highly processed input, SplitNet has a shared layer that extracts features from the input data before splitting them into separate populations. Therefore, we consider SplitNet to bear some similarities to the population coding in the cerebellum and are therefore suitable for conducting our empirical study.

We assess the potential benefits of the error-based organization of neurons in the design of ANNs under varying training conditions and assess its effect on the learned model. 
Our empirical evaluation demonstrates the effectiveness of the considered architecture in improving the generalization of the model over standard training under challenging scenarios. It provides considerable performance gains under class imbalance which is inherent in real-world datasets and significantly improves the sample efficiency of the model, enabling the model to generalize better with fewer data. 
Additionally, our empirical results suggest that error-based organization of neurons can reduce the texture bias and vulnerability to unintended shortcut cues which improve generalization to out-of-distribution data. We attribute these improvements to the flexibility of the subnets to explore the feature space more and learn specialized features for the semantic groups. Furthermore, our analyses of the characteristics of the model suggest that it compresses more information and converges to flatter minima. We would like to emphasize that all of these benefits come merely from the design of the network rather than explicit regularization or specific techniques for each scenario. Our empirical results highlight the potential of error-based grouping and partial error views based learning mechanisms in ANNs.

Our work aims to bring the perspective of population coding based design in ANNs and presents it as a promising direction for further research.
We believe that exploring the design space of population coding based ANNs can lead to more reliable and robust models that may address some of the key challenges and limitations of current AI models.

\section{Background and Methodology}
Here, we provide the required background and the premises of our study. Section \ref{cerebellum_overview} first provides an overview of population coding in the cerebellum study~\cite{shadmehr2020population} along with key insights and implications for ANNs, Section \ref{splitnet_overview} provides a detailed overview of SplitNet, which is central to our analysis, and finally, Section \ref{pop_coding_ann} draws parallels between the two and explains how we aim to study the potential benefits of population coding in ANNs.

\subsection{Population Coding in the Cerebellum} \label{cerebellum_overview}
The cerebellum is an important learning site in the brain~\cite{rapoport2000role,vandervert2016prominent} and, therefore, several studies in neuroscience have scrutinized how efficient learning is enabled in the cerebellum ~\cite{herzfeld2020principles,herzfeld2018encoding,kitazawa1998cerebellar,kojima2010changes}. It has a relatively simple circuit architecture that resembles a three-layer feedforward network of neurons in which the “hidden layer” consists of Purkinje cells (P-cells), and the output layer consists of deep cerebellar nucleus (DCN) neurons. Our study focuses on the recent work of Shadmehr \etal~\cite{shadmehr2020population} which provides an extensive overview of the learning characteristics and organization of neurons in the cerebellum from a machine learning perspective and its implications. Unlike an ANN, P-cells are grouped into small populations that converge on single DCN neurons. Furthermore, the error signal conveyed to the P-Cells, which in turn act as surrogate teachers for the downstream DCN neurons they project to, is not a fair reflection of the entire error space, but is rather biased to provide a limited (personalized) view of the error space. This error-dependent grouping of P-cells into populations is believed to play a crucial role in enabling efficient learning in the cerebellum. Our study aims to bring this perspective to ANNs and to study the potential benefits of such an architecture.

\subsection{SplitNet} \label{splitnet_overview}
To this end, we consider an existing ANN architecture that bears some resemblance to a grouping of neurons with personalized error views. 
SplitNet~\cite{kim2017splitnet} was originally proposed to optimize the inference speed of the model by learning a tree-structured network architecture that is highly parallelizable. The method involves splitting the network into a set of subnetworks that share a common lower layer and using a disjoint set of features for the specific group of classes associated with the subnetwork. SplitNet employs a two-stage learning scheme whereby in the first stage classes-to-group and features-to-group assignment matrices are learned along with the network weights while regularizing them to be disjoint across groups. The learned assignment matrices are then utilized to obtain a tree-structured network that involves no connection between branched subtrees of semantically disparate class groups, which are subsequently finetuned with the cross-entropy loss. 

Concretely, for a given number of groups, $G$, the vector of assignment of the feature group and the vector of assignment of the class group for the group $g$ ($1 \leq g \leq G$) are given by $p_g \in R^D$ and $q_g \in R^K$ where $D$ is the dimension of the features and $K$ is the number of classes. $p_g$  and $q_g$ define a group together, where $p_g$ represents the features associated with the group and $q_g$ indicates a set of classes assigned to the group. The disjoint set of classes and features are learning by imposing a constraint on the network weight at each layer $W^{l}$ to be a block-diagonal matrix, where each block $W^l_{g}$ is associated with a class group $g \in G$. There is no overlap between groups, either in features or classes, so each disjoint group of classes has exclusive features associated with it. The regularization which assigns features and classes into disjoint groups consists of three objectives:

\textit{- Group Weight Regularization}, $R_{W}$ prunes out inter-group connections to obtain block-diagonal weight matrices by minimizing the off-block-diagonal entries;
\begin{equation} \label{eq1}
\begin{split}
    R_W(W,P,Q) & =  \sum_{g} \sum_{i} \|((\mathbb{I}-P_{g})WQ_{g})_{i*}\|_2 + \sum_{g} \sum_{j} \|(P_{g}W(\mathbb{I}-Q_{g}))_{*j}\|_2
\end{split}
\end{equation}
where $P_g = diag(p_g)$ and $Q_g = diag(q_g)$ are the feature and class group assignment matrices for group $g$, and $(M)_{i*}$ and $(M)_{*j}$ denote $i$-th row and $j$-th column of $M$. Eq. \ref{eq1} imposes row/column-wise $l_{2,1}$-norm on the inter-group connections.

\textit{- Disjoint Group Assignment}, $R_{D}$ ensures that the group assignment vectors are mutually exclusive by enforcing orthogonality;
\begin{equation}
    R_{D}(P,Q) = \sum_{i<j}p_{i} \cdot p_{j} + \sum_{i<j}q_{i} \cdot q_{j}
\end{equation}
\textit{- Balanced Group Assignment}, $R_{E}$ encourages the group assignments to be uniformly distributed by minimizing the squared sum of elements in each group assignment vector.
\begin{equation}
    R_{E}(P,Q) = \sum_{g}\Big((\sum_{i} p_{gi})^2 + (\sum_{j}q_{gj})^2\Big)
\end{equation}
Therefore, the overall regularization loss is as follows;
\begin{equation}
\begin{split}
    \Omega(W,P,Q) & = \lambda_{1}R_{W}(W,P,Q) + \lambda_{2}R_{D}(P,Q) + \lambda_{3}R_{E}(P,Q)
\end{split}
\end{equation}
where $\lambda_1$, $\lambda_2$, and $\lambda_3$ control the strength of each regularization. For more details, see~\cite{kim2017splitnet}.

\subsection{Studying the potential benefits of Population Coding in ANNs} \label{pop_coding_ann}
The resemblance of the cerebellum to a feedforward network and a preliminary understanding of the error-driven organization of neurons and the learning mechanisms it employs provide us with an opportunity to study the benefits of such an architecture in ANNs. Standard learning consists of evaluating an overall error term (e.g. mean cross-entropy loss over a training batch) and subsequently updating each neuron's weight in the gradient direction, which minimizes the loss term. As explained in Section \ref{cerebellum_overview}, this is in stark contrast to how the cerebellum learns, and therefore we aim to study the potential impact of a similar error-dependent grouping of neurons into populations and subsequently learning from partial error views in ANNs.

To this end, we first define the framework within which we conduct our study by drawing parallels under the classification task in ANNs. We aim to learn semantically disparate grouping of object classes which can be represented by a disjoint set of features. Semantically similar classes are likely to share features and meaningfully partition the input space. Therefore, the classification error associated with each semantic group can provide a personalized view of the error space, which can be subsequently utilized to learn specialized features in the associated subnetwork (population of neurons). This learning paradigm naturally lends itself to tree-structured network architectures such as SplitNet. Figure \ref{fig:overview} shows the similarities between the cerebellum and structure and the learning dynamics of SplitNet (referred to as PC-ANN for emphasis). Notably, a closer look at the backpropagation of errors in SplitNet reveals an intriguing property that makes it suitable for our study as an instance of ANN architecture that bears similarity to the population coding in the cerebellum: the logits for each class in a semantic group depend only on the associated subtree (population) which therefore receives an error signal which is biased towards correcting the error associated with the semantic group (partial error). For instance, consider the scenario where we have two semantic groups: living and non-living, and the input image is of a cat. The logit values for non-living classes are provided by the associated subtree and vice versa. Hence, as the error signal for each unit depends on its involvement in the forward pass, the subtree for the living semantic group will receive an error signal biased towards correcting the error associated with the logits for living classes and similarly for the non-living subtree. Therefore, we posit that splitNet implicitly utilizes partial error views to create specialized populations of neurons. Studying the performance and characteristics of such a network enables us to gauge the potential benefits of mimicking population coding in ANNs.

\section{Experimental Setup} \label{sec:exp_setup}
To ensure a fair comparison, we compare the standard training and population coding based training paradigm under uniform experimental settings. Following Kim \etal~\cite{kim2017splitnet}, we employ WRN-16-8~\cite{zagoruyko2016wide} for both baseline (Standard-ANN) and SplitNet experiments. Unless otherwise stated, we use the following learning scheme: random horizontal flip and random crop data augmentations with reflective padding of 4 and mean standard normalization; Adam optimizer with $5e^{-4}$ weight decay; 100 epochs; the batch size of 64; and an initial learning rate of $1e^{-4}$,
decayed by a factor of 0.1 at epochs 10, 30 and 50. For SplitNet, we use a 2-way split (i.e. $G=2$) at the final linear layer. For all our experiments, we use $\lambda_1=1$, $\lambda_2=2$ and $\lambda_3=10$. For evaluation, we report the mean and one standard deviation of 3 runs with different seeds.

\begin{table}[t]
\centering
\caption{Test accuracy on different datasets. PC-ANN consistently improves the generalization of the model across datasets of varying complexity, demonstrating its versatility.}
\label{tab:gen}
\begin{small}
\begin{sc}
\begin{tabular}{l|ccc}
\toprule
& Cifar-10 & Cifar-100 & Tiny-ImageNet \\ \midrule
Baseline & 92.49 \scriptsize{$\pm$0.25} & 73.65 \scriptsize{$\pm$0.18} & 49.14 \scriptsize{$\pm$0.49} \\
PC-ANN   & \textbf{93.24} \scriptsize{$\pm$0.21} & \textbf{75.33} \scriptsize{$\pm$0.47} & \textbf{53.02} \scriptsize{$\pm$0.22}  \\
\bottomrule
\end{tabular}
\end{sc}
\end{small}
\end{table}

\section{Empirical Evaluation}
To study the potential benefits of incorporating a similar mechanism for population coding in ANN, we evaluate the characteristics and learning behavior of SplitNet in various challenging scenarios. Therefore, we refer to SplitNet as an instance of the desired population coding-based ANN (PC-ANN), the subnetworks as populations, and the classification error associated with the learned semantic grouping of classes as partial error views to emphasize our focus on studying the potential impact of a similar mechanism of population coding in ANNs as the cerebellum.

\subsection{Performance}
To test the versatility of the models, we consider multiple datasets of varying complexity. Table \ref{tab:gen} shows that PC-ANN consistently leads to generalization gains across datasets, especially in more complex datasets where both the number of classes and the interclass similarity are higher.
The results suggest that PC-ANN is capable of learning useful semantic groups and learning efficiently with partial error views. We believe that partial error views that provide a signal to the corresponding populations enable the model to explore the feature space more extensively and learn specialized features for each semantic group, which can help the model avoid the pitfalls of narrow learning~\cite{tramer2020fundamental}.

\begin{table*}[t]
\centering
\caption{Comparison of models trained under various levels of class imbalance. Note that the degree of imbalance increases as $\gamma$ reduces. PC-ANN provides consistent generalization gains over baseline under varying degrees of class imbalance, particularly for higher imbalance on more complex datasets.}
\label{tab:imb}
\begin{small}
\begin{sc}
\begin{tabular}{l|cc|cc|cc}
\toprule
 & \multicolumn{2}{c|}{Cifar-10} & \multicolumn{2}{c|}{Cifar-100} & \multicolumn{2}{c}{Tiny-ImageNet} \\ \midrule
$\gamma$ & Baseline           & PC-ANN             & Baseline           & PC-ANN             & Baseline         & PC-ANN         \\ \midrule
2               & 78.02 \scriptsize{$\pm$0.68} & \textbf{79.85} \scriptsize{$\pm$0.61} & 47.42 \scriptsize{$\pm$0.49} & \textbf{51.55} \scriptsize{$\pm$0.29} &   23.67 \scriptsize{$\pm$0.08}  & \textbf{28.46} \scriptsize{$\pm$ 0.07}  \\
1               & 74.59 \scriptsize{$\pm$0.42} & \textbf{75.84} \scriptsize{$\pm$0.85} & 36.87 \scriptsize{$\pm$0.34} & \textbf{42.35} \scriptsize{$\pm$0.45} &   17.74 \scriptsize{$\pm$0.29}  & \textbf{21.67} \scriptsize{$\pm$0.17}  \\
0.6             & 72.84 \scriptsize{$\pm$0.68} & \textbf{74.66} \scriptsize{$\pm$0.51} & 34.02 \scriptsize{$\pm$0.46} & \textbf{38.41} \scriptsize{$\pm$0.43} &   15.75 \scriptsize{$\pm$0.79}  & \textbf{19.54} \scriptsize{$\pm$0.16}  \\
0.2             & 71.68 \scriptsize{$\pm$0.24} & \textbf{73.57} \scriptsize{$\pm$0.90} & 30.43 \scriptsize{$\pm$0.81} & \textbf{35.10} \scriptsize{$\pm$0.49} &   13.57 \scriptsize{$\pm$0.42}  & \textbf{17.45} \scriptsize{$\pm$0.12}  \\\bottomrule
\end{tabular}
\end{sc}
\end{small}
\end{table*}

\subsection{Out-of-Distribution (OOD) Generalization}
A long-standing challenge for AI is its inability to generalize well to OOD data, while humans excel at generalizing to novel situations. To test whether population coding enables the model to learn more generalizable features, we consider two challenging scenarios. We first utilize the cleaned version of the DomainNet dataset~\cite{peng2019moment} that consists of data from different domains on 345 object classes. We train the models on the real domain and use the painting, clip art, sketch, and infograph domains for our OOD testing. We also consider different variants of the ImageNet dataset~\cite{deng2009imagenet}. ImageNet-R~\cite{hendrycks2021many} and ImageNet-B~\cite{hendrycks2021many} contain images from 11 different renditions and real blurry images from a subset of 100 classes of ImageNet classes, respectively. However, ImageNet-A~\cite{hendrycks2019natural} provides a dataset for naturally occurring adversarial examples. We test the models trained on Tiny-ImageNet datasets on the common subset of classes within each of these ImageNet variants for OOD evaluation.

\begin{figure}[t]
\begin{center}
    \includegraphics[width=.6\columnwidth]{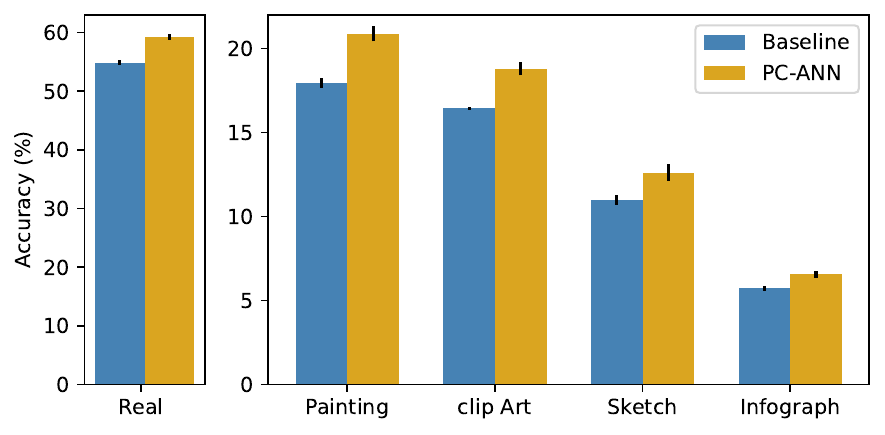}
    \caption{Generalization of models trained on real photos and tested on various out-of-distribution domains. Population coding enables the model to learn better generalizable features, leading to improved OOD generalization.}
    \label{fig:ood_domain}
    \end{center}
\end{figure}

\begin{figure}[t]
    \centering
    \begin{minipage}{0.475\textwidth}
        \centering
        \includegraphics[width=1\columnwidth]{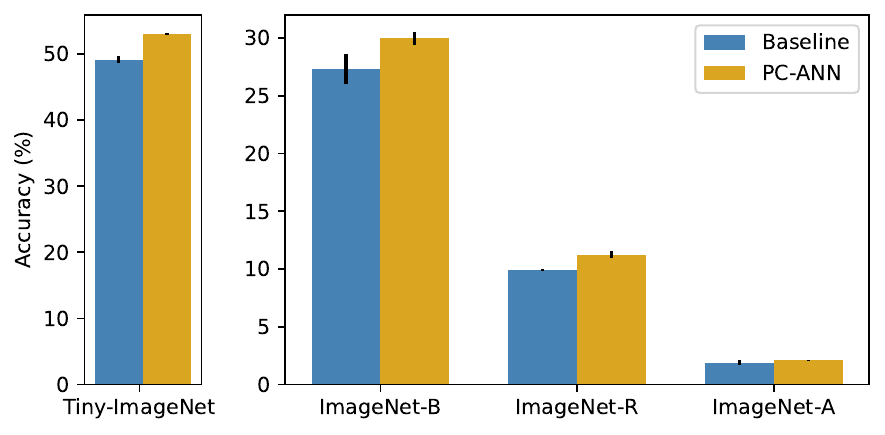}\\
        \caption{Generalization of models trained on Tiny-ImageNet and tested on common classes in different variants of ImageNet dataset. PC-ANN consistently provides better OOD generalization and shows potential for improving the adversarial robustness of the model.}
        \label{fig:ood_imagenet}
    \end{minipage}%
    \hspace{0.08in}
    \begin{minipage}{0.5\textwidth}
        \centering
        \includegraphics[width=1\columnwidth]{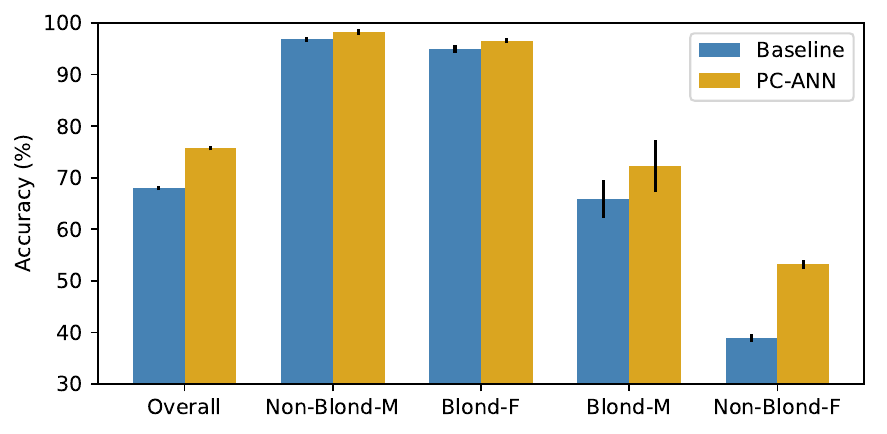}
        \caption{Shortcut learning analysis on CelebA-Skewed. PC-ANN considerably reduces the susceptibility of the model to learn the blonde color shortcut strategy as can be seen in the performance gap on Blonde-M and Non-Blond-F.}
        \label{fig:shortcut}
    \end{minipage}
\end{figure}

PC-ANN provides better generalization across all the domains of DomainNet (Figure \ref{fig:ood_domain}). We observe similar gains on different variants of the ImageNet datasets (Figure \ref{fig:ood_imagenet}). Although the difference in ImageNet-A is minor, it provides early evidence that having separate subnetworks may improve adversarial robustness. We hypothesize that the improvement in OOD generalization with PC-ANN may be attributed to learning a specialized set of features for the learned semantic groups.

\begin{table*}[t]
\centering
\caption{Performance of the models trained on a different percentage of the training samples. PC-ANN improves the sample efficiency of the model, allowing it to achieve higher performance with less amount of training data.}
\label{tab:data_eff}
\begin{small}
\begin{sc}
\begin{tabular}{l|cc|cc|cc}
\toprule
\multirow{2}{*}{\begin{tabular}[c]{@{}c@{}}Samples\\ (\%)\end{tabular}} & \multicolumn{2}{c|}{Cifar-10}        & \multicolumn{2}{c|}{Cifar-100}       & \multicolumn{2}{c}{Tiny-ImageNet} \\ \cmidrule{2-7}
           & Baseline         & PC-ANN           & Baseline         & PC-ANN           & Baseline         & PC-ANN         \\ \midrule
100        & 92.49 \scriptsize{$\pm$0.25} & \textbf{93.24} \scriptsize{$\pm$0.21} & 73.67 \scriptsize{$\pm$0.18} & \textbf{75.33} \scriptsize{$\pm$0.47} &  49.14 \scriptsize{$\pm$0.49}  &  \textbf{53.02} \scriptsize{$\pm$0.22}  \\
50         & 88.69 \scriptsize{$\pm$ 0.23} & \textbf{90.34} \scriptsize{$\pm$0.16} & 65.22 \scriptsize{$\pm$0.21} & \textbf{68.35} \scriptsize{$\pm$0.37} &  40.27 \scriptsize{$\pm$0.30}  &  \textbf{46.07} \scriptsize{$\pm$0.12}   \\
20         & 80.89 \scriptsize{$\pm$0.24} & \textbf{84.02} \scriptsize{$\pm$0.35} & 48.93 \scriptsize{$\pm$0.55} & \textbf{53.42} \scriptsize{$\pm$0.54} &  26.04 \scriptsize{$\pm$1.00}  &  \textbf{32.73} \scriptsize{$\pm$0.46}  \\
10         & 73.13 \scriptsize{$\pm$0.42} & \textbf{76.36} \scriptsize{$\pm$0.43} & 35.58 \scriptsize{$\pm$0.39} & \textbf{40.95} \scriptsize{$\pm$0.34} &  18.66 \scriptsize{$\pm$0.37}  &  \textbf{23.09} \scriptsize{$\pm$0.13}   \\
5          & 63.48 \scriptsize{$\pm$0.86} & \textbf{67.28} \scriptsize{$\pm$0.09} & 23.77 \scriptsize{$\pm$0.42} & \textbf{28.07} \scriptsize{$\pm$0.65} &  11.83 \scriptsize{$\pm$0.45}  &  \textbf{15.21} \scriptsize{$\pm$0.23}   \\
1          & 42.64 \scriptsize{$\pm$0.20} & \textbf{44.03} \scriptsize{$\pm$0.16} & ~~8.52 \scriptsize{$\pm$0.27}  & \textbf{~~9.47} \scriptsize{$\pm$0.10}  &  ~~4.35 \scriptsize{$\pm$0.23}   &  \textbf{~~4.83} \scriptsize{$\pm$0.05}    \\ \bottomrule
\end{tabular}
\end{sc}
\end{small}
\end{table*}

\subsection{Imbalanced Datasets}
The majority of the benchmark datasets have a uniform distribution of samples across the object classes. However, class imbalance is naturally inherent in the real world, where some objects are more prevalent than others, or it is relatively easier to obtain more data for certain classes. Standard training exhibits bias towards the prevalent classes at the expense of minority class~\cite{dong2018imbalanced} leading to a significant drop in generalization performance. While several approaches have been proposed for efficiently training models under class imbalance~\cite{johnson2019survey} which employs specialized techniques for tackling class imbalance or making certain assumptions about the distribution of data, we still lack a general method that improves the robustness of the underlying learning paradigm.

To evaluate the robustness of PC-ANN to class imbalance, we simulate varying degrees of class imbalance on different datasets. We follow~\cite{hendrycks2019using} and employ the power law model in which the number of training samples for a class $c$ is given by $n_c = \lfloor a/((c-1)^{-\gamma}+b) \rceil$, where $\lfloor{.}\rceil$ is the integer rounding function, $\gamma$ represents an imbalance ratio, $a$ and $b$ are offset parameters to specify the largest and smallest class sizes. The training data becomes a power-law class distribution as the imbalance ratio $\gamma$ decreases. We compare the performance of PC-ANN with the standard ANN on varying degrees of class imbalance $\gamma\in\{2.0, 1.0, 0.6, 0.20\}$ as the $\gamma$ value decreases, the class imbalance increases. ($a, b$) are set so that the maximum and minimum class counts are (5000, 250) for CIFAR-10, (500, 25) for CIFAR-100, and Tiny-ImageNet.

Table \ref{tab:imb} shows that PC-ANN consistently provides a considerable performance improvement over standard ANNs, especially for more complex datasets with high degrees of class imbalance without any explicit regularization. We believe that a major shortcoming of standard ANNs is that there is no division or specialization of neurons as each unit is involved in correcting the prediction for every input. Therefore, an imbalanced batch significantly affects the performance of the model as the entire network is adjusted to reduce the loss on the imbalanced batch, thus preferring the dominant class at the expense of less sampled classes. On the contrary, the partial error views and disjoint subtrees in PC-ANN provide more protection to parts of the network, providing implicit regularization. Furthermore, it can take the prevalence of classes into account while grouping them to mitigate the impact of dominant classes on the performance of minority classes, which builds robustness into the learning framework itself.

\subsection{Sample Efficiency}
Learning complex concepts with a few examples is a hallmark of human intelligence~\cite{zador2019critique}, whereas it remains a challenge for ANNs that are data-hungry and require an abundant amount of labeled data to generalize well~\cite{deng2009imagenet}. This limits their application in a limited data regime~\cite{yao2021machine}. We believe that mimicking the learning machinery of the brain may lead to models that can generalize better under a low data regime.
To this end, we compare the performance of the models trained on a subset of different datasets where we only use $p \in [1, 5, 10, 20, 50]$ percentage of the training dataset and test on the full test set. Table \ref{tab:data_eff} shows that PC-ANN consistently provides better generalization compared to standard ANNs, suggesting that it can learn efficiently with limited data. Notably, the performance gains are higher for complex datasets, where both the number of classes and their interclass similarities are higher.
We hypothesize that the grouping of neurons into populations allows each population to explore different regions in the feature space, enabling the model to learn more efficiently from partial error views of fewer data.

\subsection{Shortcut Learning}
Shortcuts are decision rules that perform well on standard benchmarks but fail to transfer to more challenging testing conditions, including real-world scenarios~\cite{geirhos2020shortcut}. As the models are typically trained to maximize the training accuracy, they are quite likely to rely on spurious correlations: associations that are predictive of the training data, but not valid for the actual task. A major challenge for enabling efficient learning in ANNs is therefore to control the sensitivity of the training process to such spurious correlations.
To evaluate the susceptibility of the model to shortcut learning, we follow the analysis in~\cite{jain2021combining} and consider a gender classification task based on CelebA dataset~\cite{liu2015deep} (CelebA-Skewed) where the training dataset is biased so that it only contains blonde females and non-blond males. Therefore, hair color is highly predictive on training data but not in test data where hair color and gender are independent attributes. Therefore, this may result in a decision rule based only on hair color. We use the same training scheme and only change the learning rate decay steps to 60 and 90 epochs.

\begin{figure*}[t]
    \centering
    \includegraphics[width=0.7\textwidth]{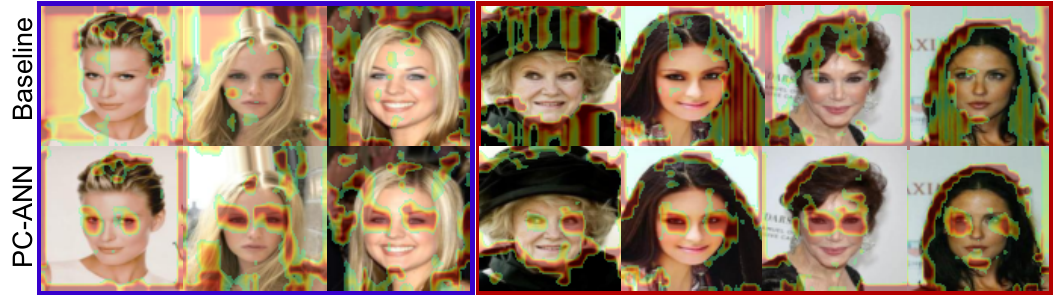}
    \caption{Visual explanations of the models trained on CelebA-Skewed. Attention maps suggest that PC-ANN relies on the salient features of the face to predict gender, while standard training relies more on the unintended shortcut cue (blond hair color in this case).}
    \label{fig:celeba-viz}
\end{figure*}

Figure \ref{fig:shortcut} shows that PC-ANN is in fact less vulnerable to shortcut learning and significantly improves generalization compared to standard ANN. Particularly, we see considerable gain in generalization to non-blond females and blond males without any explicit regularization. To better understand how the two models make decisions, we use the Grad-CAM~\cite{selvaraju2017grad} approach to examine the visual explanations of the models. We use the penultimate layer to extract the feature embeddings and use a threshold of 0.4 on the attention maps. Figure \ref{fig:celeba-viz} shows that population coding remarkably enables the model to attend to the salient features of the face to distinguish between the genders, while standard training relies more on the unintended shortcut cue of the hair color, thus paying more attention to the hair region to inform its decision. The remarkable ability of PC-ANN to avoid the shortcut cue without any explicit regularization can also be attributed to the flexibility of PC-ANN to learn specialized disjoint features for each gender and avoid narrow learning.

\begin{figure}[t]
    \centering
    \begin{minipage}{0.48\textwidth}
        \centering
        \includegraphics[trim={0 0 1.5cm 0},clip, width=1\columnwidth]{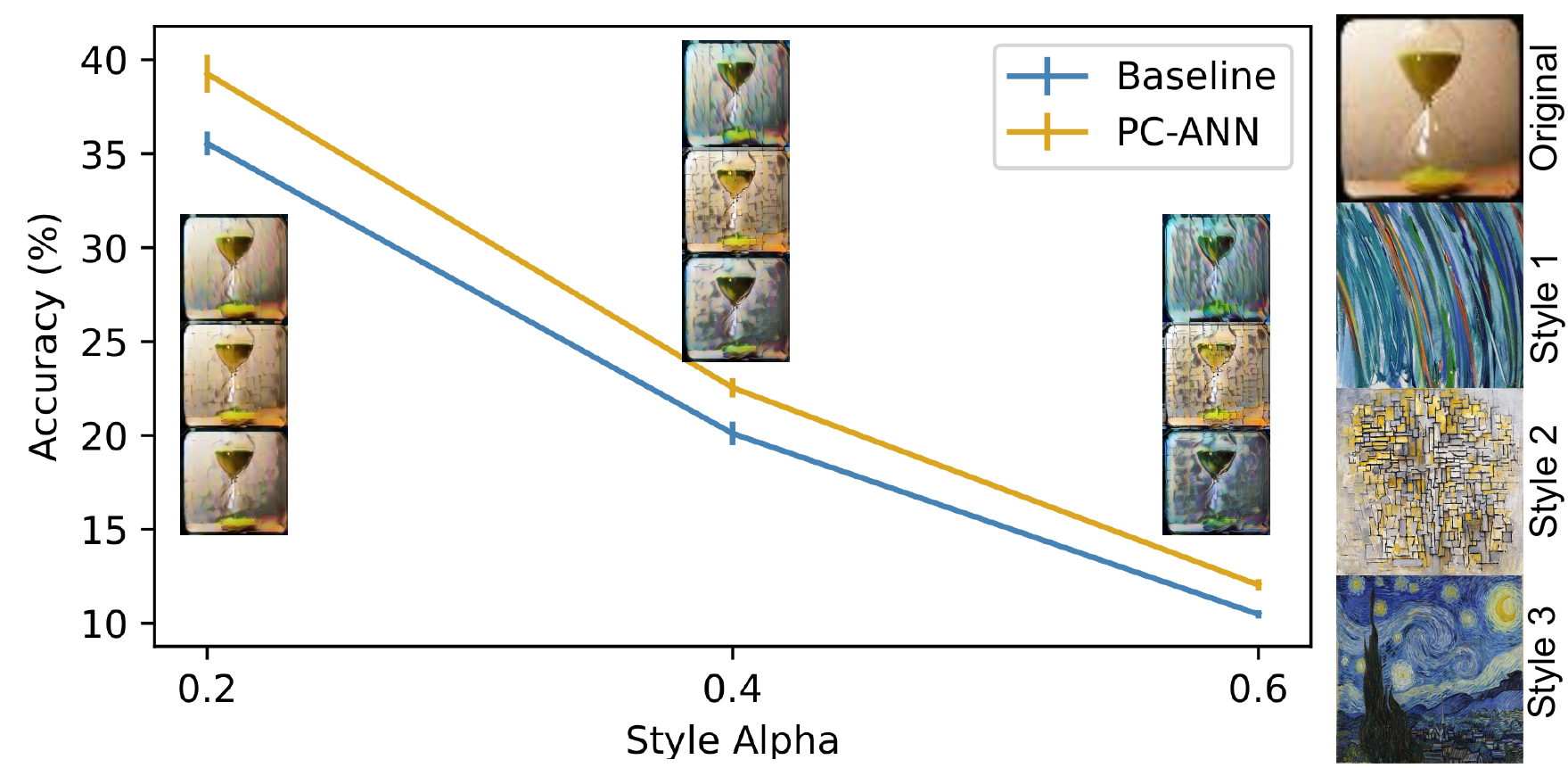}
        \caption{Texture bias analysis of models trained on Tiny-ImageNet under varying degrees of stylization. PC-ANN provides higher generalization on the stylized images, indicating lower texture bias compared to standard ANNs. The images on the right show the original images and three style images whereas the images on the graph show stylized test images at different strengths.}
        \label{fig:texture_bias}
    \end{minipage}%
    \hspace{0.08in}
    \begin{minipage}{0.48\textwidth}
        \centering
        \includegraphics[width=1\columnwidth]{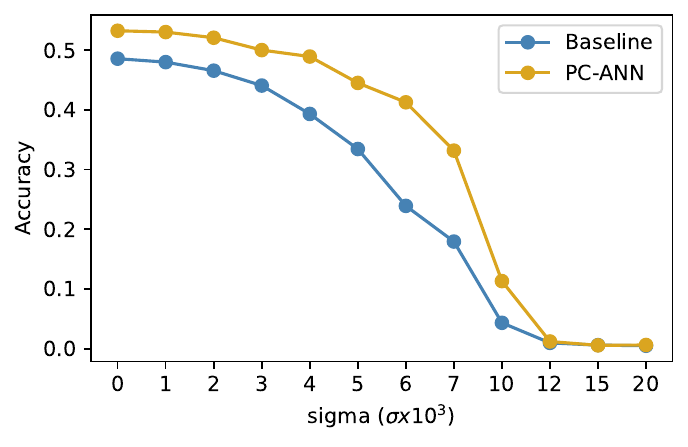}
        \caption{Training accuracy of models on Tiny-ImageNet under varying degrees of weight perturbations. PC-ANN is more stable to weight perturbations, indicating convergence to flatter minima.}
        \label{fig:tiny-perturb}
    \end{minipage}
\end{figure}


\section{Characteristics Analysis}
Here, we analyze the characteristics of PC-ANN to gain insight into generalization improvements.

\subsection{Texture Bias}
Geirhos \etal~\cite{geirhos2018imagenet} conducted a comparative study on convolutional neural networks (CNNs) and human observers on images with a texture-shape cue conflict. Their study revealed that, in sharp contrast to humans, CNNs are strongly biased towards recognizing texture instead of shape, alluding to fundamentally different classification strategies. They further showed that models that learn shape-related features are more robust and generalizable whereas models that rely on texture are susceptible to shortcut learning and result in poor generalization. As PC-ANNs bear some similarities to population coding in the cerebellum, we aim to investigate whether they exhibit behavior that is closer to humans than standard ANNs.

Following~\cite{geirhos2018imagenet}, we evaluate the texture bias of the model by applying style transfer~\cite{huang2017arbitrary} to the Tiny-ImageNet test images. We use four different style images and apply style transfer with varying strengths, i.e. style alpha $\in [0.2, 0.4, 0.6]$ so that only the shape of the image corresponds to the correct label.
Figure \ref{fig:texture_bias} shows that PC-ANN is able to generalize better under varying stylization strengths, suggesting that it is less biased toward the texture of the image.

\subsection{Convergence to Flatter Minima}
As the loss landscape of DNN's optimization objective is non-convex, there can be multiple solutions that can fit the training data, some solutions, however, generalize better because of being in wider valleys where the model predictions do not change drastically with small perturbations in the parameter space compared to the narrow crevices~\cite{selvaraju2017grad,chaudhari2019entropy,keskar2017large}. 
To assess whether PC-ANN converges to wider minima, we follow the analysis in~\cite{zhang2018deep} and add independent Gaussian noise of increasing strength to the parameters of the trained model and analyze the generalization of the trained models on the training dataset. Figure \ref{fig:tiny-perturb} shows that the performance is more stable to the perturbations, suggesting convergence to wider minima.

\subsection{Information Compression}
A number of studies that view ANNs from an information theory perspective~\cite{tishby2015deep, shwartz2017opening} relate the degree to which ANNs compress the information in their hidden states to the bounds on generalization, with higher information compression leading to a stronger bound. 
To evaluate the effect of population coding on the compression of information in the learned representation, we follow the analysis in~\cite{lamb2019interpolated} by freezing the learned representation of the model and measuring how well the frozen representations can fit random labels. we add a 2-layer multi-layer perceptron (MLP) network with 400 and 200 neurons on top of the frozen models trained on the different datasets and fit them on random binary labels. Table \ref{tab:info-compression} shows that PC-ANN enables higher information compression suggesting that the disjoint set of features in PC-ANN allows the model to learn optimal representations that can compress higher semantic information.


\begin{table}[t]
\centering
\caption{Comparative analysis on the degree to which a model with frozen learned representations can fit random binary labels. Lower training accuracy indicates higher information compression.}
\label{tab:info-compression}
\begin{small}
\begin{sc}
\begin{tabular}{l|ccc}
\toprule
& Cifar-10 & Cifar-100 & Tiny-ImageNet \\ \midrule
Baseline & 51.80 \scriptsize{$\pm$1.08} & 92.50 \scriptsize{$\pm$2.57} & 74.37 \scriptsize{$\pm$3.89} \\
PC-ANN   & \textbf{51.60} \scriptsize{$\pm$0.65} & \textbf{90.58} \scriptsize{$\pm$2.21} & \textbf{71.33} \scriptsize{$\pm$0.78}  \\
\bottomrule
\end{tabular}
\end{sc}
\end{small}
\end{table}



\section{Conclusion} \label{discussion}
We conducted an empirical study to explore the potential benefits of drawing insights from neuroscience findings to the development of AI algorithms. Here, we focused on the recent study~\cite{shadmehr2020population}, which explains the error-based organization of neurons in the cerebellum from a machine learning perspective and attempted to draw parallels with an existing tree-structured ANN. 
Our empirical evaluation of the considered architecture shows improved robustness to class imbalance and shortcut learning, efficient learning under limited data, and reduced texture bias. Furthermore, the characteristic analyses demonstrate that it compresses higher information in the hidden states, and converges to flatter minima. 
We hypothesize that these benefits are a consequence of the architecture that resembles population coding in the cerebellum, and further work to explicitly mimic the error-based grouping of neurons in ANN is a promising research direction.

\textbf{Limitations and Future work:} Our study focused on the object recognition task where a meaningful semantic grouping of the classes is possible and utilizes an existing suitable tree-based architecture. As such, the network does not explicitly mimic population coding in the cerebellum, and it is not straightforward to employ it for other tasks (e.g. regression) or when semantic grouping is not possible.
We hope that our study inspires exploration of this idea in different domains.

Some potential focus areas for future work can be better strategies for forming error-based groupings of neurons and partial error views and aligning them to minimize the global task error, intertwining the population formation and learning from partial views instead of two separate stages of learning, and explicitly biasing the update rule of the population towards the partial views while also varying the strength of the update in different layers.

{\small
\bibliographystyle{plain}
\bibliography{egbib}
}

\end{document}